\documentclass[sigconf]{acmart}

\usepackage{glossaries}
\usepackage{siunitx}
\usepackage{subcaption}

\AtBeginDocument{%
  \providecommand\BibTeX{{%
    \normalfont B\kern-0.5em{\scshape i\kern-0.25em b}\kern-0.8em\TeX}}}

\copyrightyear{2022}
\acmYear{2022}
\setcopyright{acmlicensed}\acmConference[IXR '22]{Proceedings of the 1st
Workshop on Interactive eXtended Reality}{October 14, 2022}{Lisboa, Portugal}
\acmBooktitle{Proceedings of the 1st Workshop on Interactive eXtended Reality
(IXR '22), October 14, 2022, Lisboa, Portugal}
\acmPrice{15.00}
\acmDOI{10.1145/3552483.3556458}
\acmISBN{978-1-4503-9501-4/22/10}

\newacronym{XR}{XR}{Extended Reality}
\newacronym[plural=GANs]{GAN}{GAN}{Generative Adversarial Network}
\newacronym{PCA}{PCA}{Principal Component Analysis}
\newacronym{t-SNE}{t-SNE}{t-distributed Stochastic Neighbor Embedding}
\newacronym{TSTR}{TSTR}{Train on Synthetic, Test on Real}
\newacronym{LSTM}{LSTM}{Long Short-Term Memory}
\newacronym[plural=GRUs,longplural=Gated Recurrent Units]{GRU}{GRU}{Gated Recurrent Unit}
\newacronym{PDF}{PDF}{Probability Density Function}
\newacronym{CDF}{CDF}{Cumulative Distribution Function}
\newacronym{DTW}{DTW}{Dynamic Time Warping}

\begin{document}

\title{Generating Realistic Synthetic Head Rotation Data for Extended Reality using Deep Learning}

\author{Jakob Struye}
\email{jakob.struye@uantwerpen.be}
\affiliation{%
  \institution{University of Antwerp - imec}
  \city{Antwerp}
  \country{Belgium}
}

\author{Filip Lemic}
\email{filip.lemic@upc.edu}
\affiliation{%
  \institution{Universitat Politècnica de Catalunya}
  \city{Barcelona}
  \country{Spain}
}

\author{Jeroen Famaey}
\email{jeroen.famaey@uantwerpen.be}
\affiliation{%
  \institution{University of Antwerp - imec}
  \city{Antwerp}
  \country{Belgium}
}

\renewcommand{\shortauthors}{Jakob Struye, Filip Lemic, \& Jeroen Famaey}

\begin{abstract}
  Extended Reality is a revolutionary method of delivering multimedia content to users. A large contributor to its popularity is the sense of immersion and interactivity enabled by having real-world motion reflected in the virtual experience accurately and immediately. This user motion, mainly caused by head rotations, induces several technical challenges. For instance, which content is generated and transmitted depends heavily on where the user is looking. Seamless systems, taking user motion into account proactively, will therefore require accurate predictions of upcoming rotations. Training and evaluating such predictors requires vast amounts of orientational input data, which is expensive to gather, as it requires human test subjects. A more feasible approach is to gather a modest dataset through test subjects, and then extend it to a more sizeable set using synthetic data generation methods. In this work, we present a head rotation time series generator based on TimeGAN, an extension of the well-known Generative Adversarial Network, designed specifically for generating time series. This approach is able to extend a dataset of head rotations with new samples closely matching the distribution of the measured time series.
\end{abstract}

\ccsdesc[500]{Mathematics of computing~Time series analysis}
\ccsdesc[500]{Computing methodologies~Adversarial learning}
\ccsdesc[300]{Human-centered computing~Virtual reality}

\keywords{Synthetic data, Data Generation, Extended Reality, Generative Adversarial Networks}


\maketitle

\section{Introduction}\label{sec:intro}
\Gls{XR}, encompassing Virtual, Mixed and Augmented Reality, has proven to be a major revolution in media consumption. In addition to its widespread use for recreational purposes~\cite{RiftGaming}, \gls{XR} has enabled novel approaches for other tasks, including training~\cite{XRTraining}, remote operation~\cite{XRRemote}, and architecture and construction~\cite{XRConstruction,XRConstruction2}. A key enabler of the \gls{XR} experience is how it can reflect the user's real-world motion accurately and immediately within the experience~\cite{XRTracking}. This enables the user to seamlessly and intuitively change their gaze direction, and can also serve as a source of input for virtual experiences. 

The user freedom in \gls{XR} induces a number of challenging demands on the system. When the user rotates their head, the displayed content must adapt to this at a moment's notice. More specifically, the \textit{motion-to-photon latency} dictates that the effect of any user motion must be visible on-screen within \SI{20}{\milli\second} as to avoid nauseating the user~\cite{MTP}. Several algorithms aid in fulfilling this latency requirement. Generated content is often warped right before display through algorithms such as Asynchronous Time-Warp, using the most recent measurements of user pose~\cite{ATW}. When displaying pre-recorded \SI{360}{\degree} content, viewport-dependent encoding ensures that only the content expected to be within the user's field of view is transmitted to reduce transmission latency~\cite{Viewport1,Viewport2}. Furthermore, video away from the user's expected centre of gaze may be encoded at lower quality, further reducing data size~\cite{GazeStreaming}. Overall, algorithms aiming at satisfying the motion-to-photon latency often include Deep Learning components converting users' orientational data to useful outputs, such as how to compress visual data. Training and testing these Deep Learning algorithms is a notoriously data-hungry process~\cite{DLSlow1,DLSlow2}. Furthermore, extensive evaluation of full algorithms again requires massive amounts of orientational data. 

While the aforementioned algorithms are the most ubiquitous consumers of orientational data in the field of \gls{XR}, needs for substantial orientational data sources arise in other situations as well. For truly wireless interactive \gls{XR}, where content is generated off-device and streamed in real-time over the air, frequencies in the millimetre-wave band (\SI{30}{\giga\hertz} to \SI{300}{\giga\hertz}) or higher are needed to stream content in extremely high quality~\cite{MTP,VRMMwave}. To guarantee sufficient signal strength in these frequency ranges, communication must be \textit{beamformed} between the sender and receiver, rather than being sent and received omnidirectionally~\cite{beamform}. In addition, in the field of Redirected Walking, non-deterministic mappings between real-world motion and virtual motion are applied to avoid real-world collisions without restricting virtual freedom~\cite{RW}. High-performance solutions in both of these cases again require enormous amounts of orientational data for training and evaluation.

Over the past years, many datasets of \gls{XR} orientational measurements have been published. Commonly, these datasets consist of logs of timestamped orientations, measured at a regular interval and represented in the yaw-pitch-roll format. In this format, an orientation is deconstructed into three subsequent rotations starting from a reference orientation: yaw represents turning one's head left or right, pitch represents tilting up or down, and roll represents tilting sideways. Collecting these datasets usually involves tens to hundreds of test subjects, who are each shown several minutes to several tens of minutes of \gls{XR} content~\cite{Corbillon,Lo,Li,Wu,AVtrack,Nasrabadi,Xu,Hu,Chakareski}. Clearly, gathering these datasets is an expensive and labour-intensive process that does not scale well. Therefore, a clearly more efficient approach is to apply synthetic data generation techniques to augment existing datasets with new, unique samples, without changing the distribution of the overall dataset~\cite{Augmentation}. Despite this, research into this approach has so far been very limited, with only an exploratory work proposing data generation through Fourier transforms~\cite{FFTmodel}. This approach considers time series of orientations as signals, which are converted to power spectral densities, after which the \textit{mean} power spectral density is modeled. Then, perturbed versions of this model are converted back to signals and finally orientational time series. This however results in synthetic time series that closely match the \textit{mean} of the set of input time series, rather than their full \textit{distribution}. In contrast, we propose to use a significantly more capable method of synthetic data generation, namely the \gls{GAN}~\cite{GAN}. A \gls{GAN} consists of two sub-systems trained in parallel. A \textit{generator} generates synthetic samples, while a \textit{discriminator} attempts to classify samples as real or synthetic. In a zero-sum game, both sub-systems interactively improve their performance: the discriminator discovers features indicative of synthetic samples, while the generator learns to avoid introducing such features. Ideally, the generator eventually outputs unique samples indistinguishable from the real ones. As each sample within a dataset of orientational data is a \textit{time series}, credible synthetic samples must not only match the original distribution when observing individual time steps, but also when observing their evolution through time. A modification to \glspl{GAN} called \textit{TimeGAN} aims to satisfy this requirement~\cite{TimeGAN}. Hence, in this work, we rely on TimeGAN to generate realistic synthetic orientational data samples. During training, the TimeGAN is provided sequences of orientational data, such that it eventually learns to generate similar, but previously unseen sequences. We repeat this process with multiple datasets to show the approach works generally. In this work, we only apply TimeGAN to orientational data, and not positional data, as many applications, including beamforming, viewport-dependent encoding and redirected walking, rely mostly on orientational data. Positional data, being significantly less dynamic, has only a limited impact in these applications~\cite{covrage, shadow}.

To gauge the utility of these synthetic datasets, one needs a metric of how similar the distributions of the respectively real and synthetic datasets are. Several general-purpose metrics are commonly used for this purpose, such as \gls{PCA}~\cite{PCA}, \gls{t-SNE}~\cite{tsne} and \gls{TSTR}~\cite{RCGAN}. These metrics are however all difficult to interpret intuitively. It is unclear when these metrics indicate a ``realistic'' synthetic dataset in absolute terms, meaning their main use is in comparing different sources of synthetic data. Fortunately, the orientational data considered in this work is, by itself, easily interpreted intuitively. As such, we opt to forego the more generally used metrics described above, and instead define a number of metrics specific to head rotation data, which together characterise the important features of the dataset. Specifically, or set of metrics considers not only the distribution of orientations, but also how often and how smoothly users rotate. We consider this to be a more convincing indication of our approach's effectiveness.

Our work contains the following contributions:
\begin{itemize}
  \item We present the first Deep Learning approach for generating realistic synthetic datasets of \gls{XR} head rotations capable of extending any dataset with minimal expert input.
  \item We outline a number of metrics intuitively characterising such datasets, allowing for interpretable estimation of the practical similarity of real and synthetic datasets.
  \item We train the model on two different datasets, and evaluate the output using the above metrics, showing that it can generate realistic datasets, outperforming the current state of the art.
\end{itemize}

The remainder of this paper is structured as follows. Section~\ref{sec:rw} covers related work. In Section~\ref{sec:method}, we outline our approach and how to quantify its performance. This performance is then evaluated in Section~\ref{sec:evaluation}. Finally, Section~\ref{sec:conclusions} concludes this work.
\section{Related Work}\label{sec:rw}
While the intersection of orientational data collection and synthetic data generation is still in its infancy, the two fields separately are well-developed. This section provides an overview of the two.
\subsection{XR pose datasets}\label{sec:rwdata}
Over the years, a wide array of datasets containing orientations or poses (i.e., locations plus orientations) of \gls{XR} users have been made available to the community. In this overview, we only consider works presenting novel datasets (i.e., not compiled from previous works) at a reasonably high sampling rate, which are, at the time of writing, readily available online.

Corbillon \textit{et al.} present an orientational dataset sampled at \SI{45}{\hertz} gathered from 59 subjects shown 6 minutes of \SI{360}{\degree} video~\cite{Corbillon}. Lo \textit{et al.} collected a \SI{30}{\hertz} dataset containing orientations along with saliency maps, identifying objects that attract attention, using 10 minutes of \SI{360}{\degree} video shown to 50 subjects~\cite{Lo}. The set of videos contains both recorded and pre-generated content, further subdivided into slow-paced and fast-paced content. Li \textit{et al.} showed 221 test subjects a 1 to 2 minute sequence of videos, repeated 10 times, gathering orientation and gaze direction at 48 to \SI{60}{Hz}~\cite{Li}. They then used this as input to a Deep Learning model predicting perceived visual quality. Next, a dataset by Wu \textit{et al.} contains the full pose at \SI{100}{\hertz} for 48 users exposed to nearly 90 minutes of video~\cite{Wu}. For the AVtrack orientational dataset, Fremerey \textit{et al.} showed 10 minutes of video to 48 subjects~\cite{AVtrack}. With only \SI{10}{\hertz} and an angular precision of \SI{1}{\degree}, this dataset is tailored more towards investigating longer-term behaviour of subjects. In Nasrabadi \textit{et al.}'s dataset, orientation was measured at \SI{60}{\hertz} from 14 minutes of video shown to 60 subjects~\cite{Nasrabadi}. Xu \textit{et al.} showed 35 minutes of video to 58 subjects, recording orientation and gaze at \SI{60}{\hertz}~\cite{Xu}. Next, Hu \textit{et al.} recorded orientation and gaze at \SI{100}{\hertz} with 30 subjects~\cite{Hu}. Each was shown 7.5 minutes of video four times, each time with a separate task, then a neural network was trained to classify the measurements according to the performed task. Zerman \textit{et al.} showed one to a few minutes of volumetric video to 20 test subjects, logging location and orientation at \SI{55}{\hertz}~\cite{Zerman}. Subramanyam \textit{et al.} showed four point cloud videos to 26 subjects for varying lengths of time, recording location and orientation at \SI{30}{\hertz}~\cite{Subramanyam}. Finally, contrary to the previous datasets, Chakareski \textit{et al.} used a navigable virtual experience in which three test subjects were allowed to navigate freely for six 2 minute sessions, during which the full pose was gathered at \SI{250}{\hertz}~\cite{Chakareski}.
\begin{figure*}[t]
  \centering
  \includegraphics[width=0.9\linewidth]{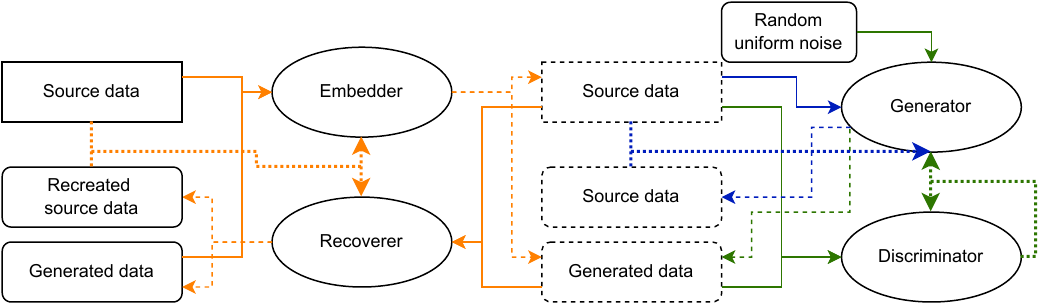}
  \caption{Illustration of the TimeGAN training process, displaying neural networks (ellipses) and data (rectangles). Rectangles with rounded corners indicate data generated by the model, while those with sharp corners represent data taken from input sequences. Each line is part of training the latent space (orange), unsupervised adversarial training (green) or supervised training (blue). Solid lines are data inputs, dashed lines are data outputs and dotted lines are loss signals. Losses originating from two data sets are a dissimilarity measure (e.g., mean squared error) while the loss originating from the discriminator is a classification loss (e.g., cross-entropy).}
  \label{fig:timegan}
\end{figure*}

Overall, while datasets vary in terms of number of subjects, sampling rate and sample length, most employ pre-recorded or pre-generated video, as opposed to an interactive environment. Datasets with truly rapid motion (e.g., from a fast-paced video game) are, to the best of our knowledge, currently non-existent. These are crucial for several applications. For example, real-time beamforming becomes inherently more challenging under rapid motion, meaning that such datasets are needed to evaluate beamforming solutions in a worst-case environment~\cite{covrage,shadow}.
\subsection{Time Series Generation}
Once a reasonable amount of data is gathered using test subjects, generating synthetic, but realistic data may be necessary to obtain a sufficiently large dataset for some application. 
For this, some approaches have been proposed. 
One such approach is the classical Smith's algorithm, designed for generating instantiations of a wireless channel model~\cite{SmithAlgo,SmithAlgo2}. In essence, it considers the time series to be generated as a set of signals. Known samples are converted to frequency space using the Fourier transform. Then, random noise sequences are weighted using filter coefficients, resulting in frequency coefficients similar to those of the known samples. Converting back to the time domain using the Inverse Fourier transform results in realistic synthetic signals. Recently, Blandino \textit{et al.} generated synthetic head rotation traces using this approach~\cite{FFTmodel}. However, they only considered the \textit{mean} power spectral density in the model, meaning the distribution \textit{between} samples is lost. We will compare our solution to this Fourier-based solution using the authors' publicly available code~\footnote{https://github.com/usnistgov/vrHeadRotation}.

Data generation has also seen significant attention from the Deep Learning community, where the \gls{GAN} is generally considered to be the prime candidate~\cite{GAN}. Recently, Martin \textit{et al.} applied this approach to scanpath (i.e., a sequence of gaze directions) generation, a field adjacent to head rotation generation~\cite{Martin}. As a regular \gls{GAN} is not time series-aware, the authors added compatibility by using \gls{DTW}, a measure for similarity between time series, as a loss function. Other approaches for inserting general time series compatibility into a \gls{GAN} model have been proposed~\cite{CRNNGAN,RCGAN}, with current state of the art being TimeGAN~\cite{TimeGAN}. These approaches are covered extensively in the next section.
\section{Methodology}\label{sec:method}
This section outlines our methodology for generating realistic synthetic orientational time series using TimeGAN. We will illustrate our approach using the dataset from \cite{Chakareski}, but note that our approach is intended to be general. We first generally present the TimeGAN algorithm, and then apply it to head rotation data generation specifically.
\subsection{TimeGAN}\label{sec:timegan}
Classical approaches to data generation are often model-based, meaning that transferring the generator from one source dataset to another requires expert input, and may necessitate significant changes in design. As such, more general, model-free approaches are desirable. \glspl{GAN} are generally considered to be the prime candidate for this~\cite{GAN}. A \gls{GAN} is a general design of a Deep Learning agent, where two sub-systems interact adversarially, to eventually generate samples matching the distribution of some source dataset. The \textit{generator} receives random noise as inputs, and converts these to synthetic samples of the desired dimensions. The \textit{discriminator} is a classifier, which, through supervised training, learns to classify samples as real (i.e., from the source dataset) or fake (i.e., from the generator). While the generator, which cannot access the source dataset, will initially output essentially random noise, it is given access to the loss function of the discriminator, meaning it can adapt its output to maximise that loss (i.e., make it more difficult for the discriminator to distinguish between fake and real). Interactively, the discriminator pushes the generator to produce more realistic samples, in turn encouraging the discriminator to discover more subtle differences between real and fake. The exact implementation of the two sub-systems depends on the type of data to generate. 

In arguably the most well-known application of \glspl{GAN}, generating fake images, these are constructed using Convolutional Neural Networks~\cite{StyleGAN}. When samples are time series however, extra care must be taken to ensure that the time-correlation between different time points \textit{within} a time series is maintained. To this end, TimeGAN introduces significant augmentations to the \gls{GAN} system~\cite{TimeGAN}. Specifically, TimeGAN's discriminator and generator consist of \glspl{GRU}, a component capable of considering time-dependencies. Furthermore, \textit{embedder} and \textit{recovery} sub-systems are added, which respectively encode a time series to a latent space of lower dimension, and decode this latent space back to the original space. These two are trained first, and then the generator produces samples in this latent space, to be converted to time series through recovery. This reduces the complexity of the generator's task to a more practically feasible level. Finally, a supervised learning step is introduced. In this step, the generator is made to complete (latent representations of) incomplete time series from the source dataset. A loss function, measuring the distance between the generated time steps and the actual time steps from the source dataset, further encourages the generator to learn the time-correlation within time series. Regular adversarial learning and this supervised learning are performed alternatingly, and their loss functions are implemented as cross-entropy loss and mean squared error, respectively. The system is summarised in Figure ~\ref{fig:timegan}. TimeGAN was shown to outperform earlier \gls{GAN}-based approaches on financial and energy datasets in its initial presentation, and has since been applied successfully to medical data~\cite{TimeGANMedical}. In this paper, we use the TimeGAN approach based on the initial authors' code\footnote{https://github.com/jsyoon0823/TimeGAN}.
\subsection{Data Inspection and Preparation}
\begin{figure*}
  \begin{minipage}[t]{0.33\textwidth}
    \includegraphics[width=\linewidth]{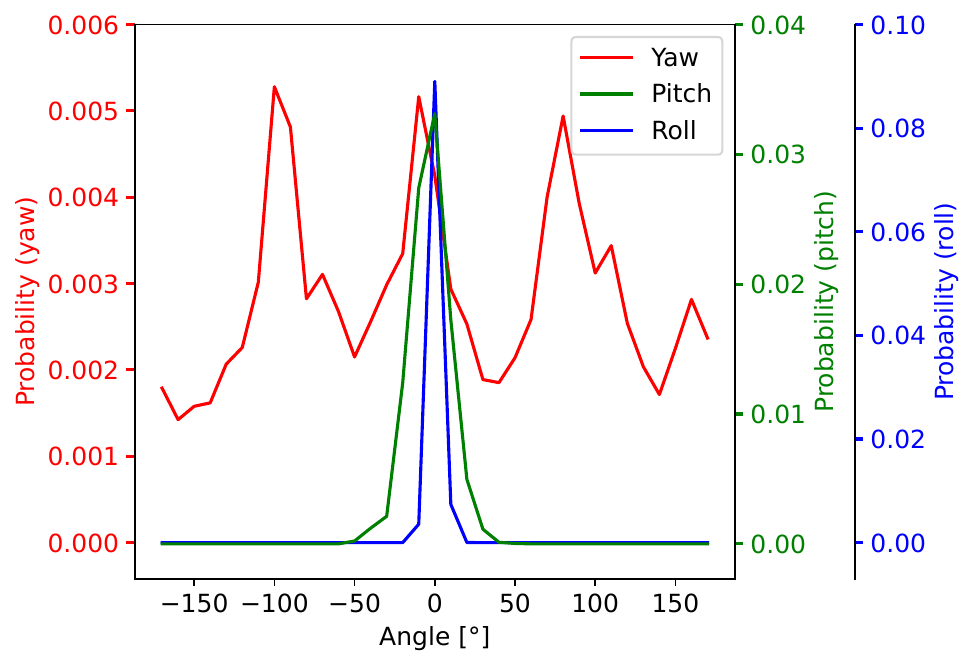}
    \caption{Distribution (quantized) of yaw, pitch and roll within the original dataset. Pitch is, by definition, restricted to $[-90,90]$ degrees.}
    \label{fig:yprdist}
  \end{minipage}%
  \hfill 
  \begin{minipage}[t]{0.31\textwidth}
    \includegraphics[width=\linewidth]{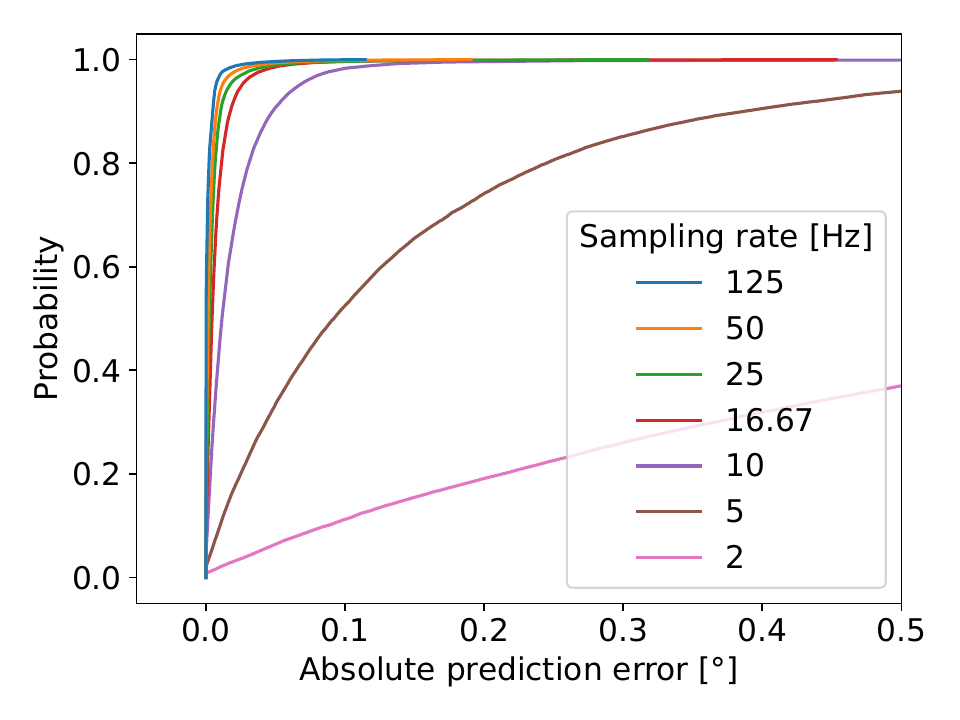}
    \caption{CDF of the absolute error between the original dataset, and the dataset first downsampled, then upsampled using a cubic spline interpolator. Errors over \SI{0.5}{\degree} are removed for clarity.}
  \label{fig:sampling}
  \end{minipage}%
  \hfill
  \begin{minipage}[t]{0.32\textwidth}
    \includegraphics[width=\linewidth]{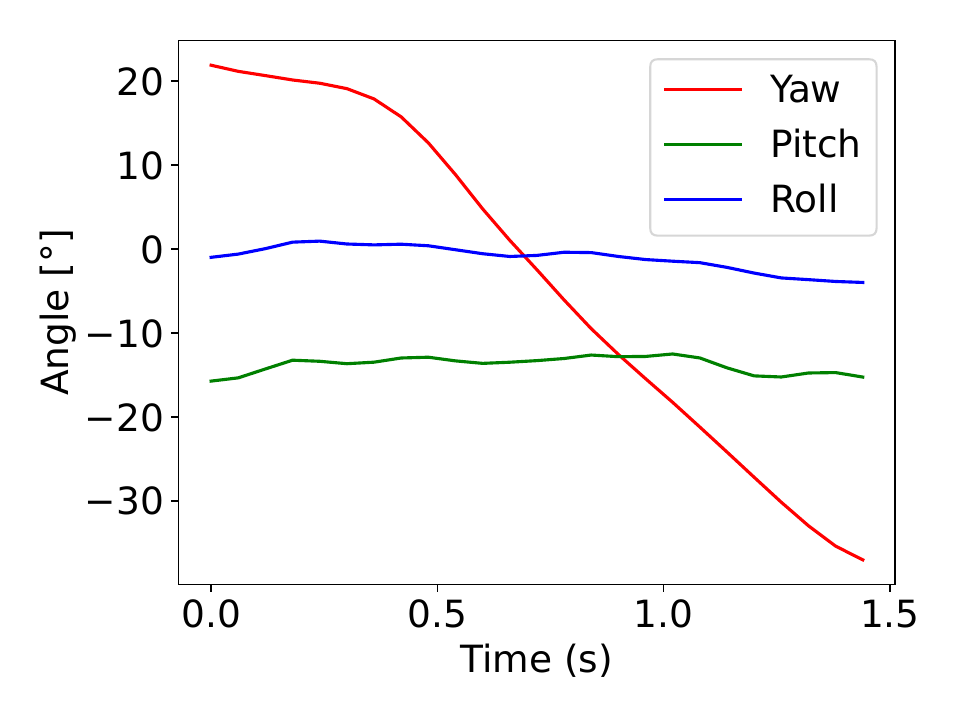}
    \caption{One sample extracted from the original dataset, after downsampling.}
  \label{fig:sample}
  \end{minipage}%
  \end{figure*}

We mainly evaluate our approach using the dataset in \cite{Chakareski}. We select this dataset to enable direct comparison with \cite{FFTmodel}, based on the same dataset. As this iteration of our work focuses on orientational data, we disregard the positional data in the dataset entirely. Remember that the dataset contains 18 orientational traces of 2 minutes each, performed by 3 test subjects, sampled at \SI{250}{\hertz}, provided in yaw-pitch-roll format. We maintain this representation as it is easily interpretable. Figure~\ref{fig:yprdist} shows the \gls{PDF} of the yaw, pitch and roll, quantized into buckets \SI{10}{\degree} wide, with the middle bucket centered around \SI{0}{\degree}. Clearly, yaw motion is significantly more pronounced. This is unsurprising, as points of interest in a virtual world are usually distributed in roughly a horizontal plane around the user. Furthermore, turning around one's axis or looking to the side is more comfortable than looking up/down or tilting one's head. Because the virtual experience was an indoor environment, users' gaze was generally aimed at one of the walls, explaining the local maxima around \SI{-90}{\degree}, \SI{0}{\degree} and \SI{90}{\degree}. While the normal distributions of the pitch and roll are easily generated by neural networks, the yaw's distribution may be more challenging on two fronts: its multiple peaks, along with discontinuities through time, whenever the representation rolls over between \SI{180}{\degree} and \SI{180}{\degree}. To combat the data's non-normality, we transform the data non-linearly using a quantile transformer, forcing a normal distribution~\cite{Quantile1,Quantile2}. Experimentation showed this to be more effective than power transforms such as Box-Cox~\cite{BoxCox} and Yeo-Johnson~\cite{YeoJohnson} for this application. To resolve the discontinuities, we simply shift the remainder of a time series by \SI{360}{\degree} whenever a discontinuity occurs. While this means we can no longer ascertain the exact data range a priori, data does remain well within reasonable bounds in practice. We emphasise that all transformations are reversible, and that synthetic data is transformed to the original representation before comparison with the source dataset.

Next, we discuss how to arrange the time series into an appropriate format for synthetic generation. Originally, the data is divided into eighteen time series of 2 minutes at \SI{250}{\hertz}, resulting in \SI{30000}{} samples per time series. As the \gls{GAN}'s training time and difficulty scale with the input size, and Deep Learning traditionally requires many distinct samples, subdividing the samples is inevitable. In line with experiments in the original TimeGAN paper, we propose to subdivide each time series into smaller instances of 25 samples each, using a sliding window. In addition, we downsample the data. We argue that downsampling does not significantly reduce the utility of the dataset. Intuitively, a person can only perform a few distinct head rotations per second. Furthermore, the law of inertia implies that these motions will be relatively smooth, and therefore easily recreated accurately through simple interpolation techniques. For a more rigorous justification, we first refer to the power spectral density estimations of the data, presented in \cite{FFTmodel}. Energy is focused around the lowest frequencies, with over \SI{90}{\percent} at \SI{5}{\hertz} and lower. As per the Shannon-Nyquist sampling theorem, this information will be maintained when downsampling to \SI{10}{\hertz}. Additionally, we investigate empirically how well the original data can be recreated after downsampling. We downsample the full dataset for different downsampling factors, then attempt to upsample back to the original frequency using a simple cubic spline interpolator. Figure~\ref{fig:sampling} shows how well interpolation performs. Up to a downsampling factor of 25 (i.e., \SI{10}{\hertz}), the difference between the actual and interpolated values remains minimal, but increases rapidly with further downsampling. Based on the analysis above, we decide to downsample the dataset to \SI{16.67}{\hertz}, a downsampling factor of 15. This results in separate time series of 1.5 seconds each. Figure~\ref{fig:sample} shows one such sample, arbitrarily selected. We expect samples of this length to be sufficient for most applications, as prediction horizons for dynamic encoding and beamforming are usually in the order of \SI{100}{\milli\second}~\cite{ViewportShort}. For cases where longer samples are desirable, one could fuse several samples together. We leave this for future work.

\subsection{TimeGAN Tuning}
The dataset, subdivided into \SI{23700}{} 1.5 second samples of 25 time points each, is provided to the TimeGAN, where the quantile transformer is fit. For proper operation, the TimeGAN's hyperparameters have to be optimised. Analogously to TimeGAN's initial evaluation, we opt to use the same network topology for every sub-system in the TimeGAN. Table~\ref{tab:hyper} summarises a well-performing set of hyperparameters. Note that we report \textit{epochs} as number of full passes through the dataset, while the reference implementation uses \textit{iterations}, each iteration being a single batch sampled from the dataset. Finally, we generate a large synthetic dataset, 10 times the size of the original, every 10 epochs. As \glspl{GAN} are notoriously challenging to train, and are known to degrade when trained for too long, this allows us to obtain a well-distributed result without having to fully optimise the epoch hyperparameter, or repeating the computationally expensive training process several times.
\begin{figure*}[t]
  \centering
  \includegraphics[width=0.32\linewidth]{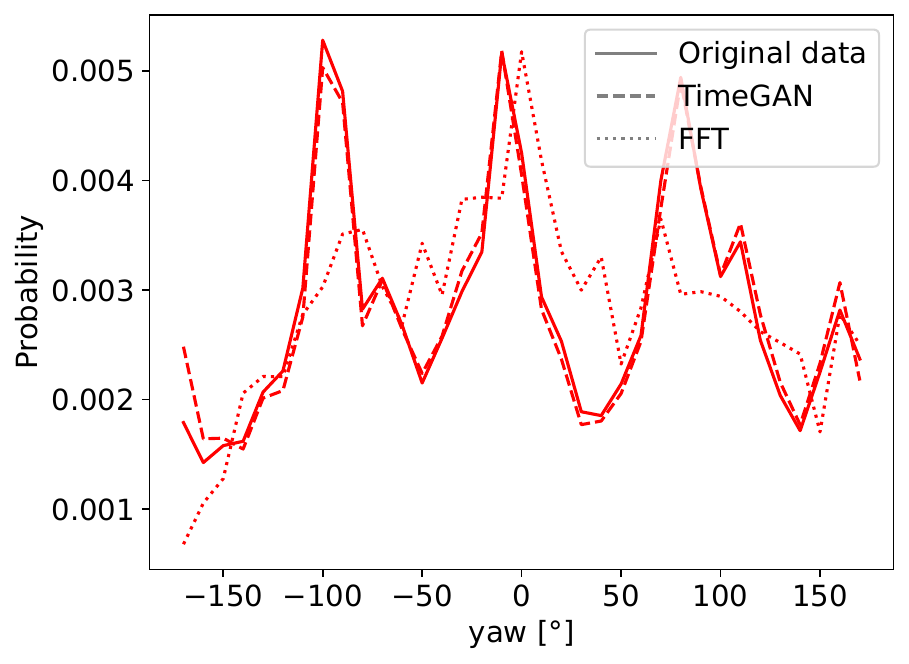}
  \hfill
  \includegraphics[width=0.32\linewidth]{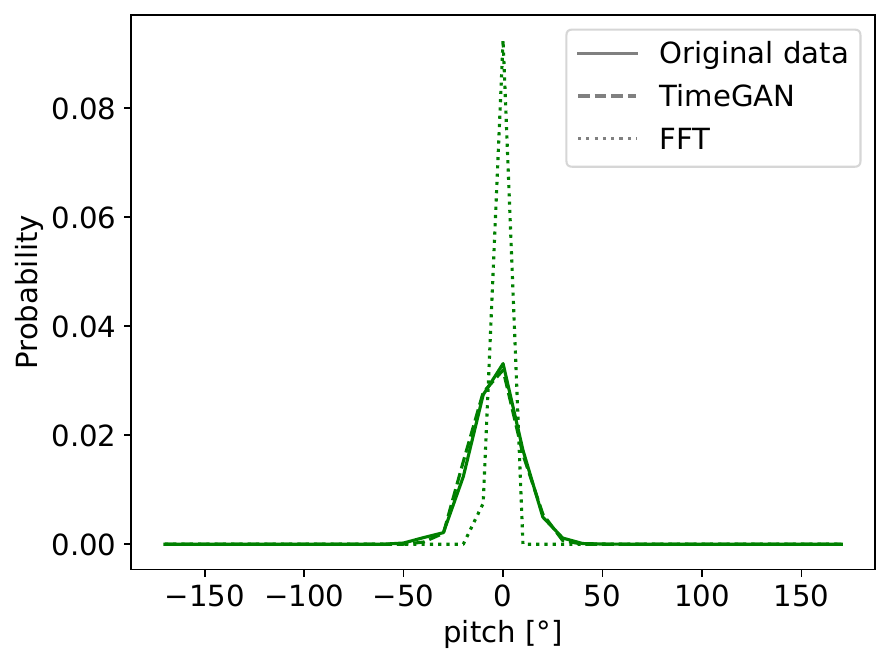}
  \hfill
  \includegraphics[width=0.32\linewidth]{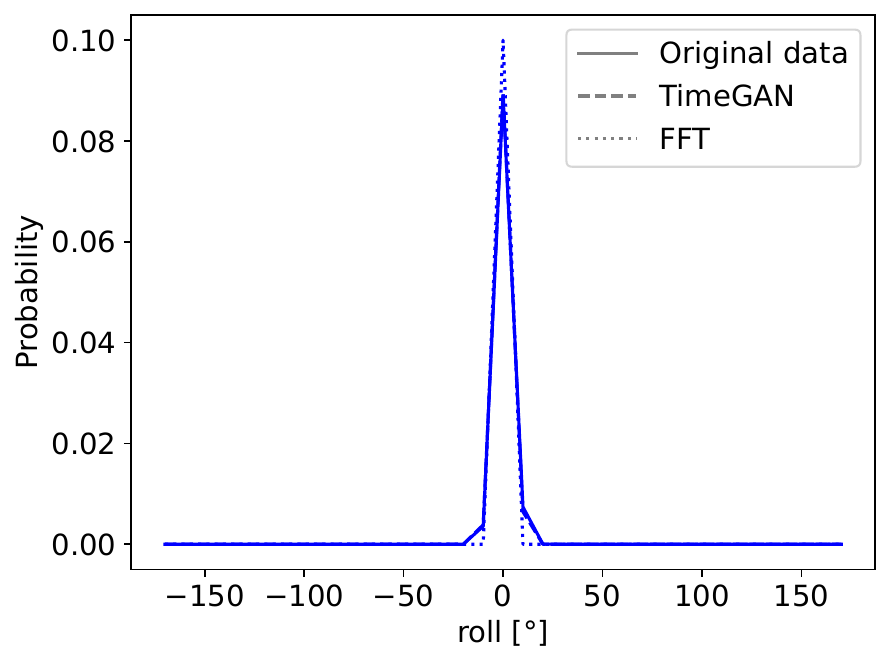}

  \caption{Distribution of yaw, pitch and roll values across all time steps of all samples}
  \label{fig:datadist}
\end{figure*}
\begin{figure*}[t]
  \centering
  \includegraphics[width=0.32\linewidth]{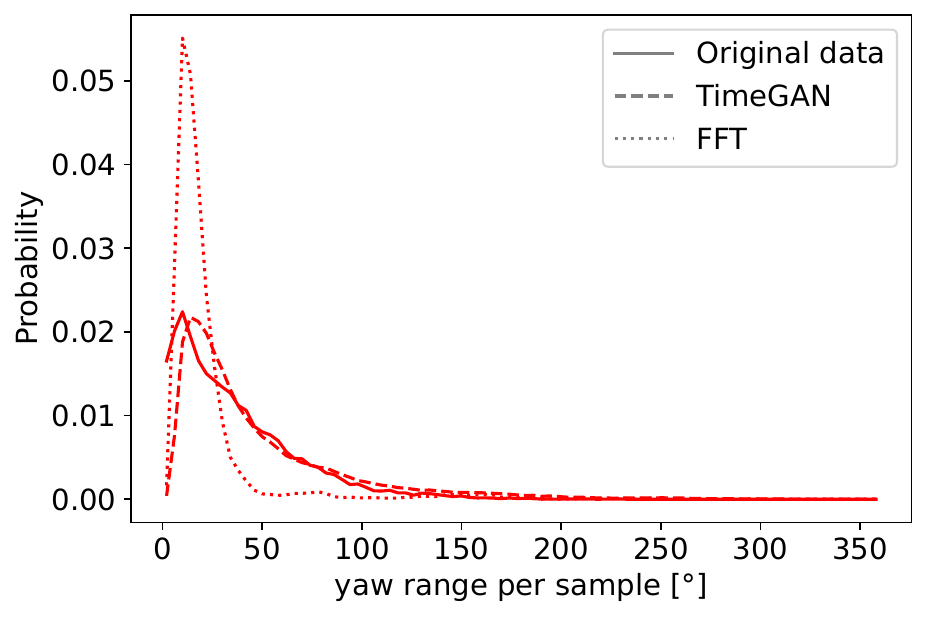}
  \hfill
  \includegraphics[width=0.32\linewidth]{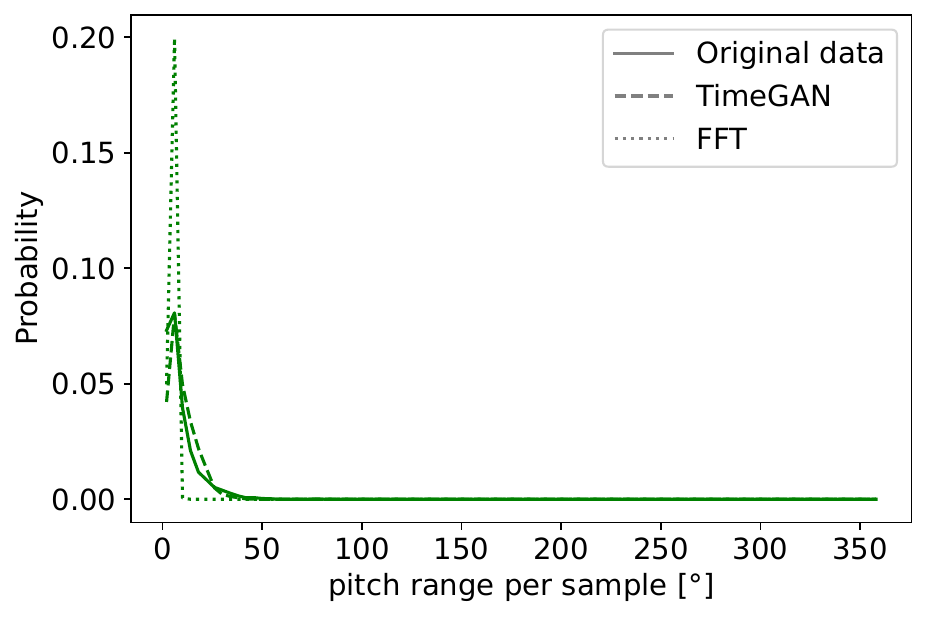}
  \hfill
  \includegraphics[width=0.32\linewidth]{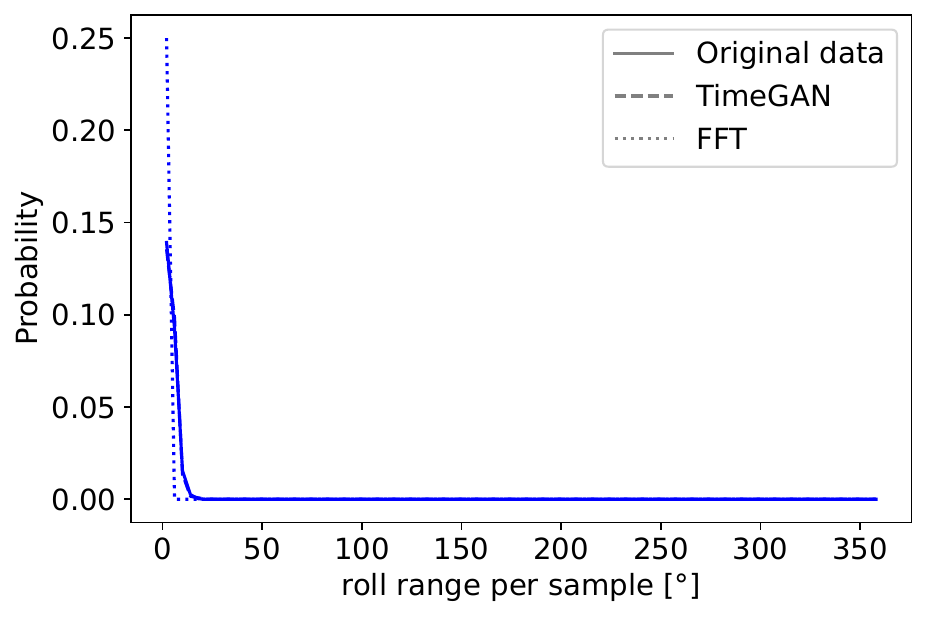}
  \caption{Distribution of the range of yaw, pitch and roll values per sample. Each range is calculated as the difference between the minimum and maximum value within the sample.}
  \label{fig:rangedist}
\end{figure*}
\subsection{Evaluation Metrics}
Once a synthetic dataset is generated, some way of evaluating how well its distribution matches the original dataset's is needed. Determining this both accurately and interpretably is notoriously difficult. Some commonly used approaches are dimensionality reduction through \glsfirst{t-SNE} or \glsfirst{PCA}, enabling visual comparison. Another is \glsfirst{TSTR}, where some neural network taking input data of the type under consideration is trained using only synthetic data, then evaluated using real data. With these approaches it is however difficult to determine what qualifies as ``good enough'', making them primarily useful when comparing data generators. For this case specifically, the dataset itself is fortunately easily interpretable. Therefore, we determine a number of metrics which together interpretably characterise the relevant features of the dataset. We propose the following metrics:
\begin{table}[t]
  \caption{TimeGAN Hyperparameters}
  \label{tab:hyper}
  \begin{tabular}{r|l}
    \toprule
    Epochs & 1250\\
    Batch size & 128\\
    Learning rate & 0.001\\
    Neurons/layer & 18\\
    Layers & 3\\
    \bottomrule
\end{tabular}
\end{table}
\begin{itemize}
  \item {\textbf{Orientation distribution}}: separate \glspl{PDF} of the yaw, pitch and roll will show whether the distribution between viewing directions is maintained. We consider every time point within one sample as a separate data point, rather than taking the mean of each sample, as to avoid masking information for discontinuous samples.
  \item {\textbf{Per-sample motion distribution}}: the distribution of the range (i.e., difference between maximum and minimum value) of yaw, pitch and roll shows whether time-correlation is maintained properly. The distribution of slower-moving and more rapid samples should ideally be maintained.
  \item {\textbf{Autocorrelation of velocity}}: The autocorrelation of the velocity (i.e., first derivative) of yaw, pitch or roll reveals whether the ``smoothness'' of the motion is maintained. 
  \item {\textbf{Cross-correlation of velocities}}: The cross-correlation of velocities quantifies time-correlation of motion between different axes. Intuitively, ``diagonal'' motion (i.e., not on only one axis) should cause such time-correlation. Ideally, this should be maintained in synthetic data, but can only be expected to be maintained if the three features are not generated independently.
\end{itemize}
We will evaluate our approach using these metrics. In addition, we also generate \gls{PCA} and \gls{t-SNE} plots, to investigate whether differences in synthetic dataset quality revealed by the above metrics are also visible in these plots.
\section{Evaluation}\label{sec:evaluation}
To obtain a synthetic dataset using TimeGAN, we trained the system for the full 1250 epochs, generating a dataset every tenth epoch. We manually inspected each result using the described metrics and selected the best-performing option. We stress that this should not be considered as overfitting. When training a network for, say, prediction, such an approach is undesirable. This would overfit the predictor to the evaluation dataset, which does not necessarily lead to a well-performing predictor for new data once the system is deployed. In this case however, the generated dataset \textit{is} the final result of the system, and the system will not be presented with inputs of (slightly) different distributions in the wild, as the input is, by definition, uniformly distributed noise.

As a baseline, we also generate a synthetic dataset with the FFT-based approach from \cite{FFTmodel}. This generates time series of \SI{30000}{} time steps, analogous to the original data, which we then downsample and subdivide into shorter time series using a sliding window.
\subsection{Head rotation metrics}
We now evaluate each of the novel metrics. Note that, when plotting distributions, values are quantized into buckets \SI{10}{\degree} wide.
\subsubsection{Orientation distribution} We first analyse the distribution of the raw yaw, pitch and roll values. We consider every time step of every sample as a separate data point and plot their distribution in Figure \ref{fig:datadist}. The distribution of roll values is rather limited, and both synthetic datasets match it closely. Roll motion requires tilting one's head, which is rather uncomfortable. The pitch's distribution is slightly wider, and here the FFT fails to match the distribution closely, while the TimeGAN again closely matches the distribution. With the complicated yaw distribution, this occurs to an even greater extent. The three peaks are closely matched by the TimeGAN, while they are significantly less pronounced for the FFT, despite only the latter being hand-crafted to match this distribution.
\begin{figure*}[t]
  \centering
  \includegraphics[width=0.32\linewidth]{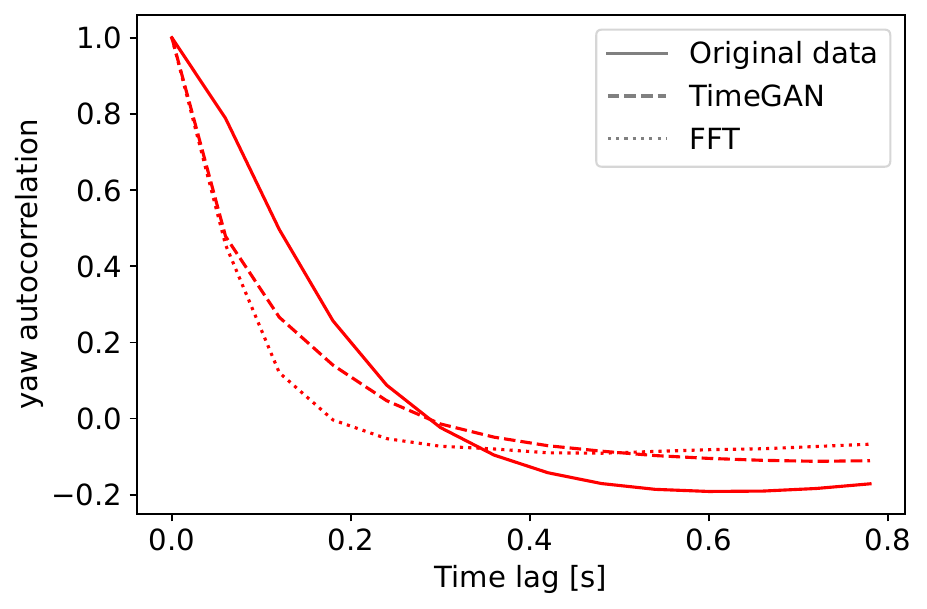}
  \hfill
  \includegraphics[width=0.32\linewidth]{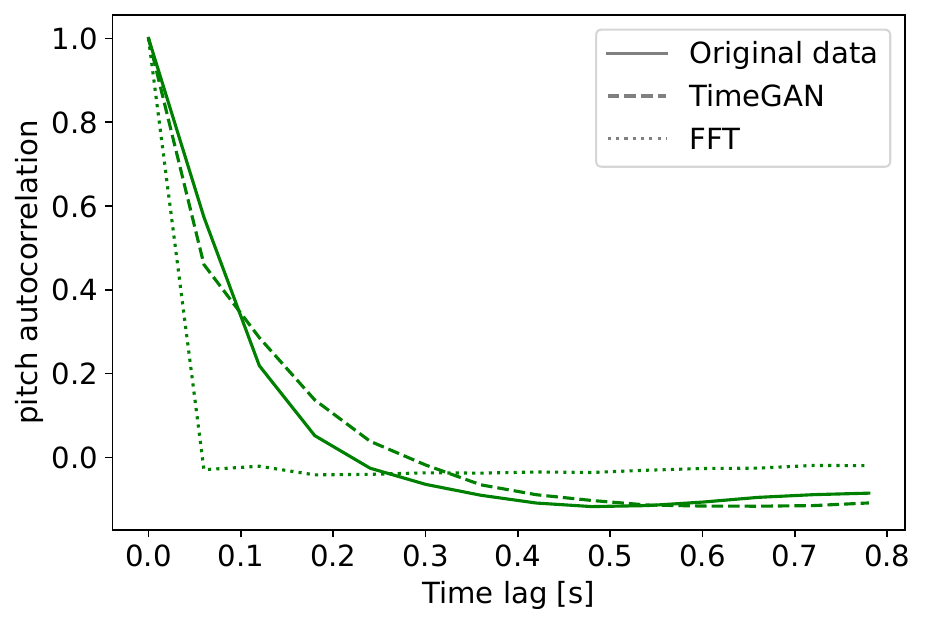}
  \hfill
  \includegraphics[width=0.32\linewidth]{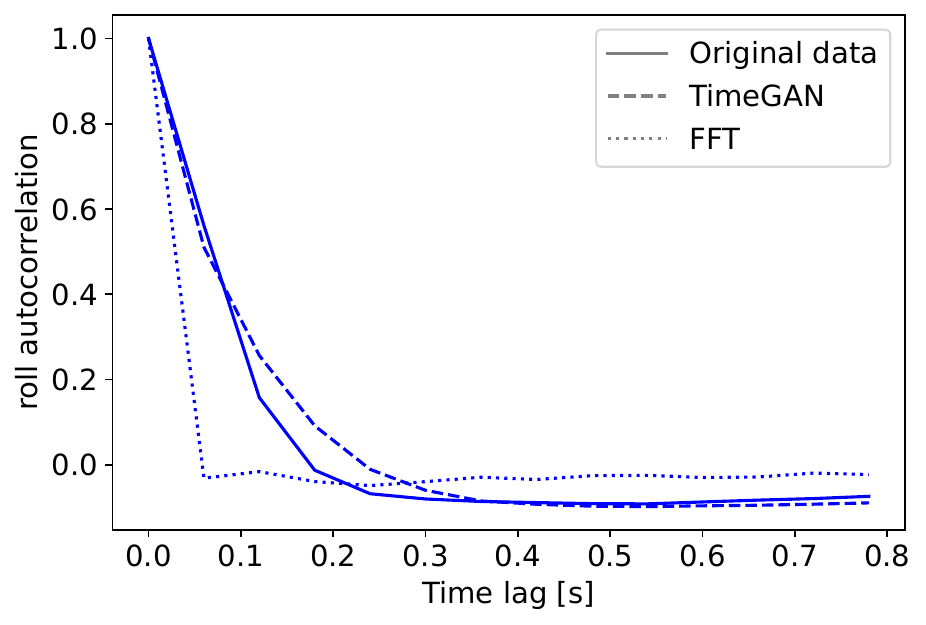}
  \caption{Mean autocorrelation of yaw, pitch and roll across all samples.}
  \label{fig:autocor}
\end{figure*}
\begin{figure*}[t]
  \centering
  \includegraphics[width=0.32\linewidth]{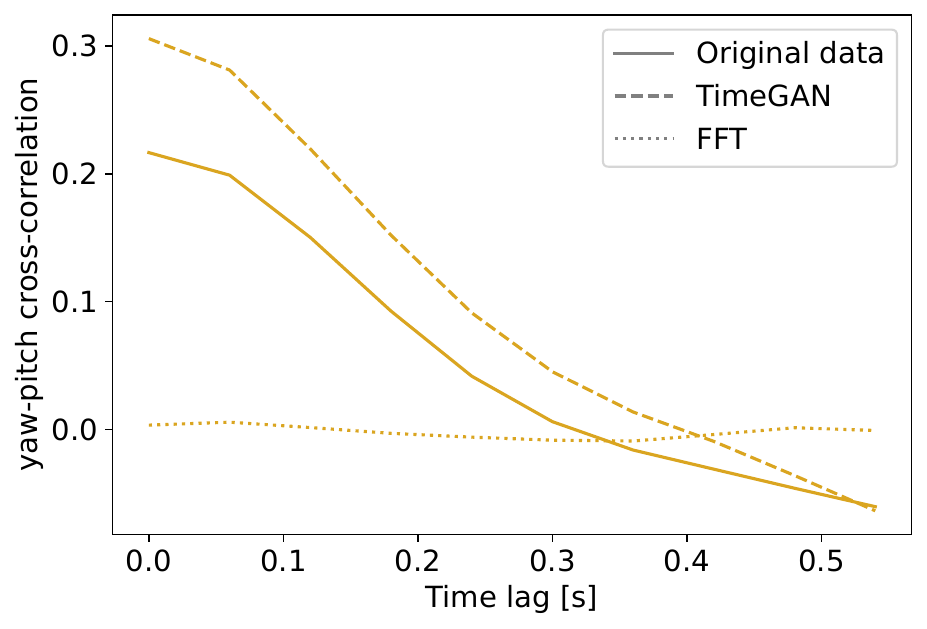}
  \hfill
  \includegraphics[width=0.32\linewidth]{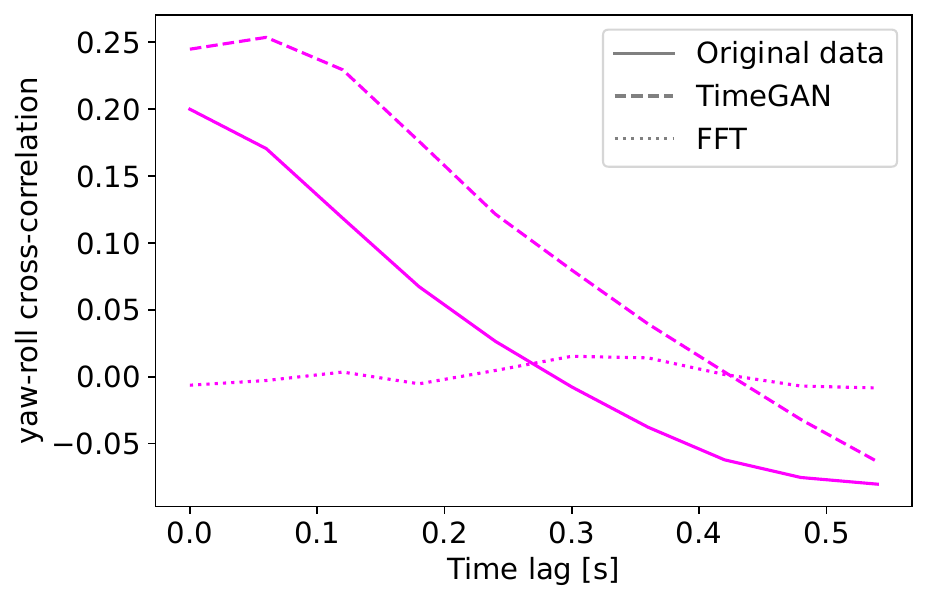}
  \hfill
  \includegraphics[width=0.32\linewidth]{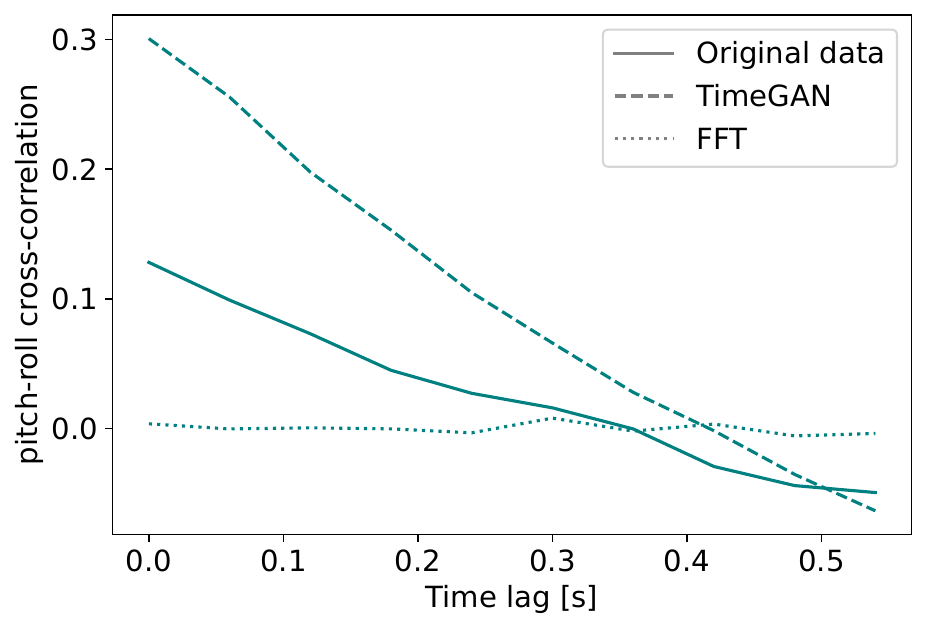}

  \caption{Mean cross-correlation of each combination of two features across all samples.}
  \label{fig:xcor}
\end{figure*}
\subsubsection{Motion distribution} Figure \ref{fig:rangedist} illustrates how well the motion of the synthetic dataset matches that of the original dataset. The FFT shows a major peak at very low motion, while the TimeGAN again closely matches the original dataset. We do note that, when considering yaw, \textit{very-low-motion} samples are underrepresented in the TimeGAN dataset. We hypothesise that, when the motion distribution is broad, generating these very-low-motion time series is inherently challenging with the TimeGAN. The time series is generated per-step, and to form a very-low-motion time series, every next step's value must be very close to the previous. A noteworthy deviation for even a single time step increases the overall motion in the entire time series. Even if the probability for generating values close to the previous ones is reasonably high, the probability of this happening for \textit{every} time step in the 25-step sequence will be low. Further evidence for this hypothesis is that when motion is consistently low, the TimeGAN easily generates this motion distribution accurately. Overall, we argue that this phenomenon is inevitable with the current approach, and leave potential solutions for future work. We do note that very-low-motion samples are less valuable. Applications such as viewport-dependent encoding or beamforming are mainly challenging under higher motion. We argue that failure to generate higher-motion samples, as occurs with FFT, is significantly more damaging to the data's utility. This failure occurs at least partially due to the FFT being designed to consistently match the distribution of the \textit{mean} time series, rather than the distribution of all time series.

\subsubsection{Autocorrelation of velocity} In a realistic time series, velocity changes gradually, largely due to the law of inertia. Intuitively, the autocorrelation, a measure of similarity between a time series and a time-lagged copy of itself, should decrease gradually as the time lag increases. This has been observed in previous work~\cite{ViewportShort}, and Figure \ref{fig:autocor} also clearly shows this behaviour for the original dataset. With the TimeGAN, the autocorrelation for pitch and roll is matched closely. With yaw, it decreases somewhat more rapidly, which is likely a side-effect of the lack of very-low-motion samples discussed above. For the FFT, the autocorrelation decreases significantly more rapidly, becoming near-zero earlier. One may expect autocorrelation to be high with the FFT, as the generated signals are of low frequency, making them inherently relatively smooth. We hypothesise that this result is at least in part due to the lower overall motion in that dataset, meaning the effect of minor perturbations is more noticeable in the (normalised) autocorrelation.

\subsubsection{Cross-correlation of velocity} Intuitively, one would expect the yaw, pitch and roll to display at least some correlation, as humans do not naturally only rotate their heads along one axis at a time. Any ``diagonal'' motion results in motion on multiple axes, while holding one's head still results in no motion on each axis. Indeed, Figure \ref{fig:xcor} confirms this intuition: the high correlation at 0 time lags shows that high motion along one axis implies a high likelihood of high motion along another. As the number of time lags increases, the correlation subsides: motion along one axis has little effect on the motion along another axis half a second later. This holds for both the original dataset and the TimeGAN, where the cross-correlation is even slightly higher. With the FFT however, this behaviour does not occur at all. The cross-correlation is near-zero for every time lag, meaning that there is no correlation between the different axes. This is unsurprising, as the FFT generates data for the three axes entirely independently from each other. Clearly, this results in a significantly less realistic dataset.
\begin{figure*}[t]
  \begin{subfigure}{0.48\linewidth}
  \centering
  \includegraphics[width=0.48\linewidth]{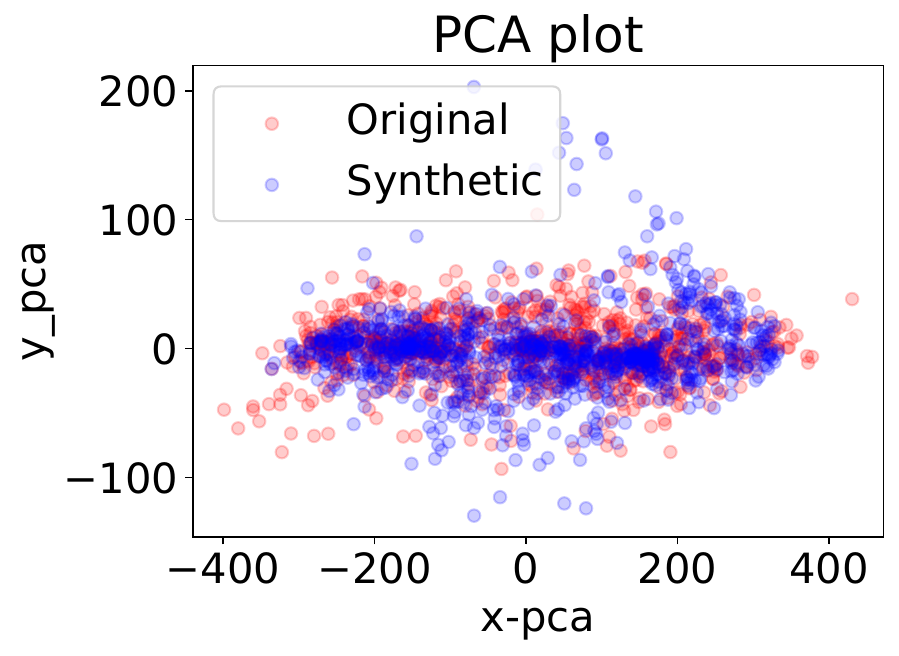}
  \hfill
  \includegraphics[width=0.48\linewidth]{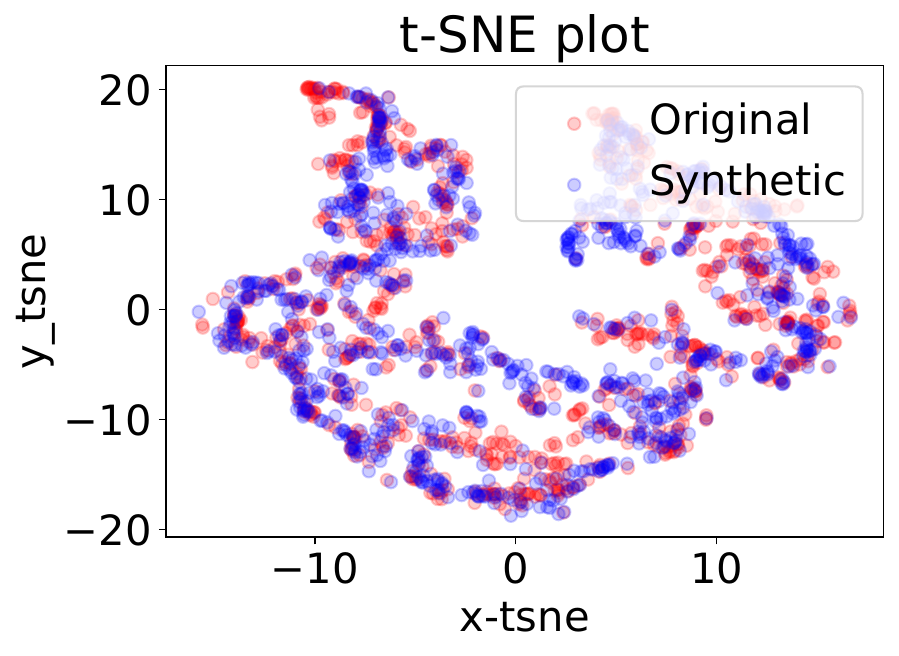}
  \caption{TimeGAN-based synthetic dataset}
  \end{subfigure}
  \hfill
  \begin{subfigure}{0.48\linewidth}
    \includegraphics[width=0.48\linewidth]{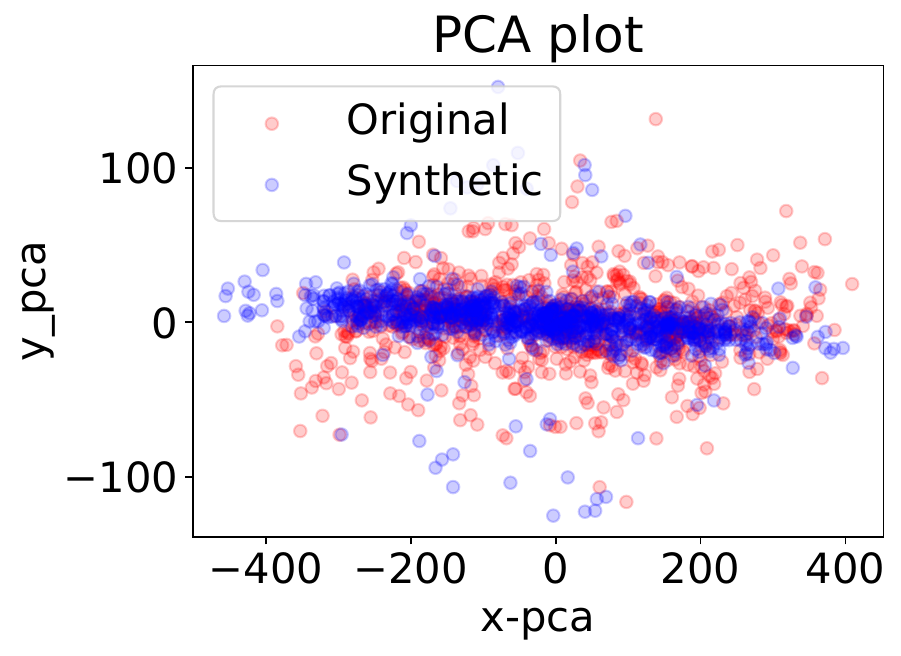}
    \hfill
    \includegraphics[width=0.48\linewidth]{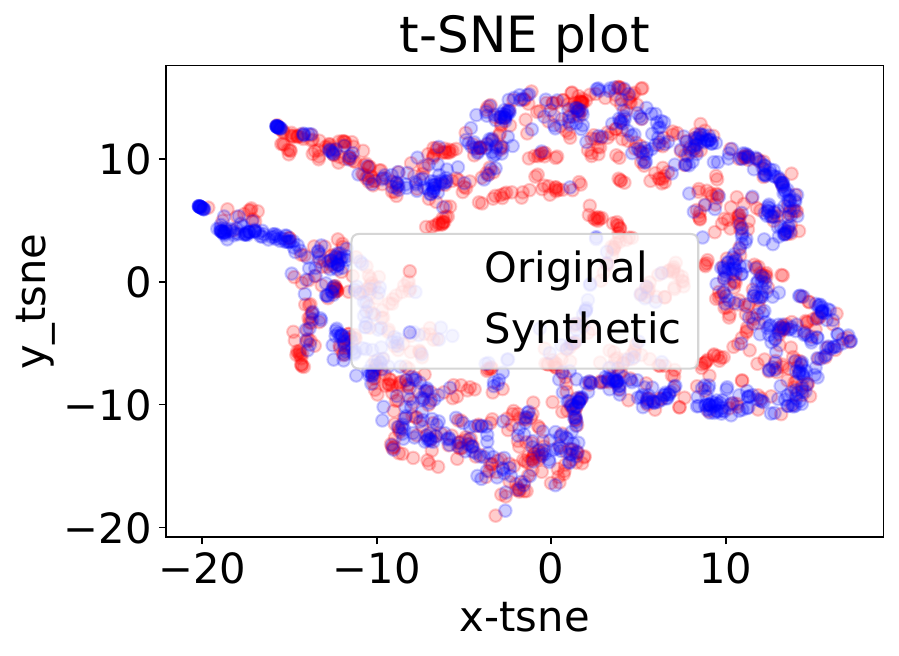}
    \caption{FFT-based synthetic dataset}
  \end{subfigure}
  \caption{PCA and t-SNE plots, comparing the distributions of the original dataset and a synthetic dataset.}
  \label{fig:pcatsne}
\end{figure*}
\begin{figure*}[t]
  \begin{subfigure}{0.24\linewidth}
    \centering
    \includegraphics[width=\linewidth]{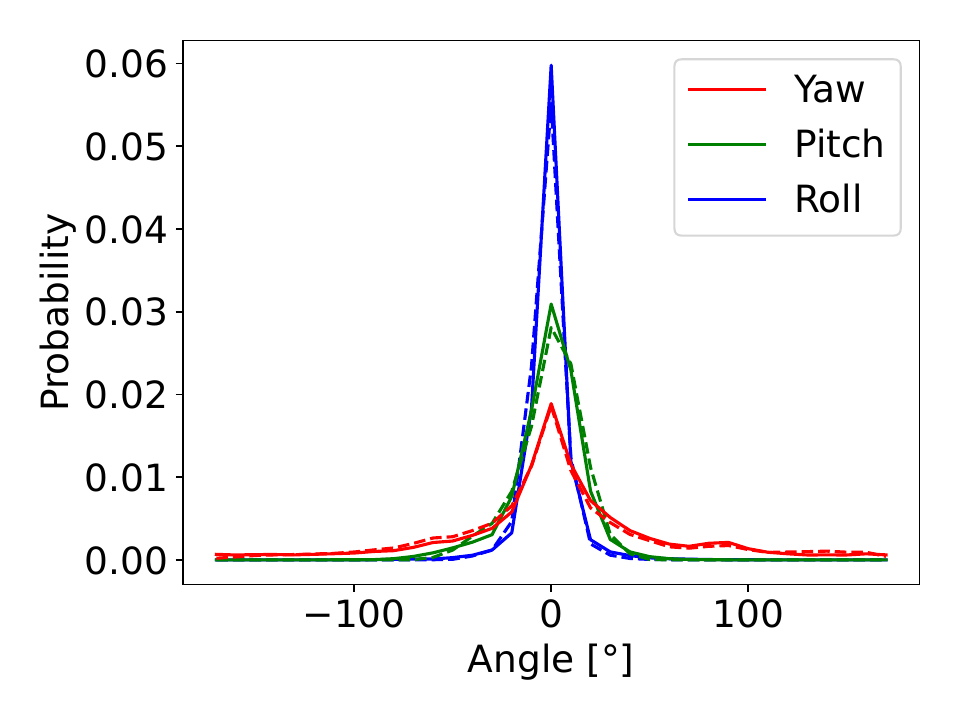}
    \caption{Orientation distribution}
  \end{subfigure}
  \hfill
  \begin{subfigure}{0.24\linewidth}
    \centering
    \includegraphics[width=\linewidth]{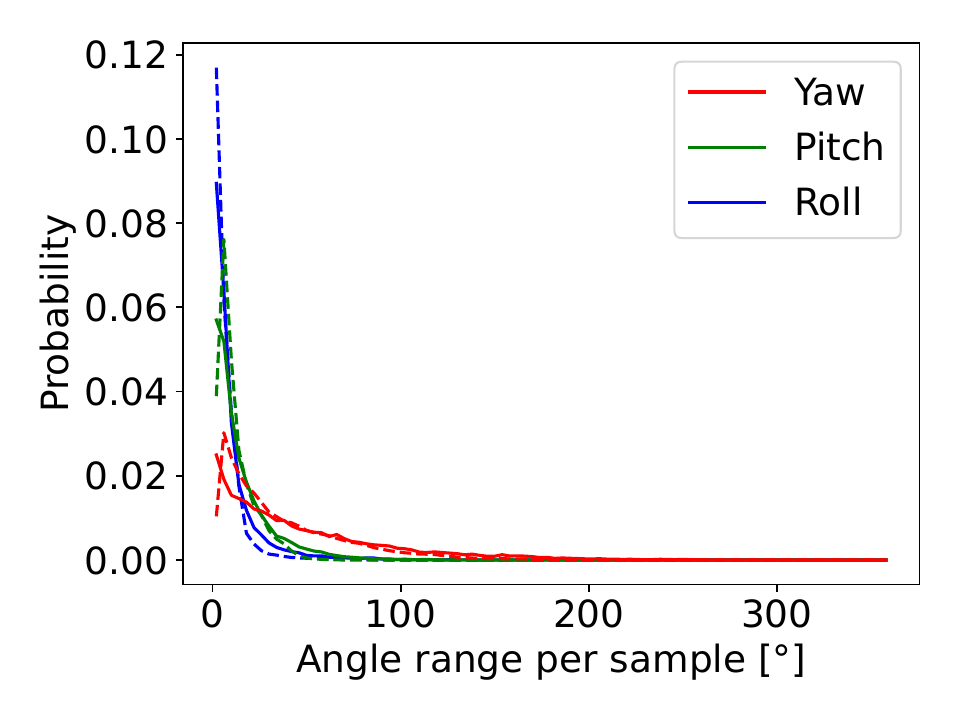}
    \caption{Motion distribution}
  \end{subfigure}
  \hfill
  \begin{subfigure}{0.24\linewidth}
    \centering
    \includegraphics[width=\linewidth]{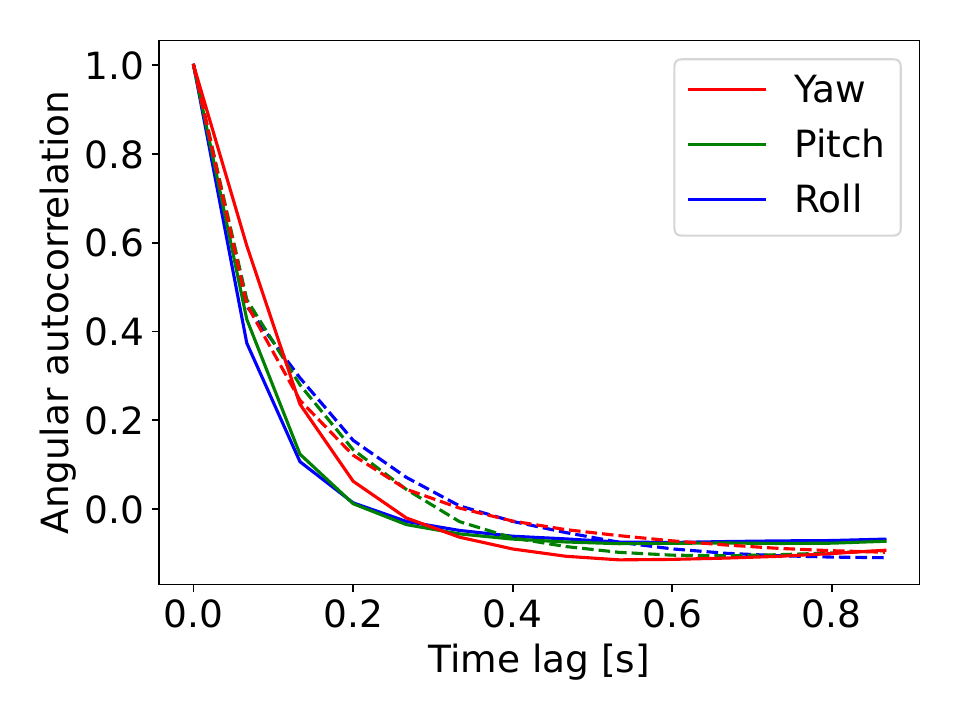}
    \caption{Autocorrelation}
  \end{subfigure}
  \hfill
  \begin{subfigure}{0.24\linewidth}
    \centering
    \includegraphics[width=\linewidth]{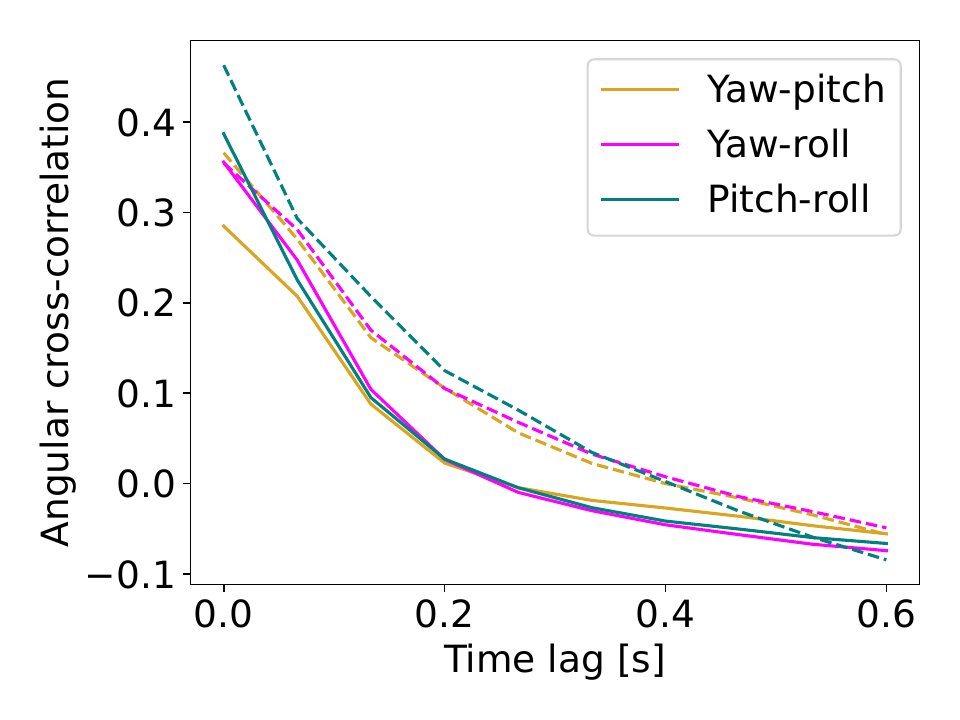}
    \caption{Cross-correlation}
  \end{subfigure}
  \caption{All metrics evaluated on a second dataset.}
  \label{fig:bonus}
\end{figure*}
\subsubsection{\gls{t-SNE} and \gls{PCA}} In addition to the head rotation metrics, we also report the commonly used \gls{t-SNE} and \gls{PCA} plots, for visual inspection of synthetic data quality. Figure \ref{fig:pcatsne} shows these for the TimeGAN and the FFT. Overlap in the plots is said to indicate similarity of the corresponding datasets. From the plots, it is difficult to tell if either synthetic dataset is superior. Despite this, our metrics clearly indicate the superiority of the TimeGAN dataset, and highlight several shortcomings of the FFT dataset. As such, we argue that this analysis does not sufficiently illustrate how realistic time series data is, proving the value of the metrics defined in this work.
\subsection{Generality}\label{sec:generality}
While we have provided some intuition for the generality (i.e., dataset-independence) of this approach, we provide further evidence by repeating the evaluation with another dataset. We selected the dataset by Lo \textit{et al.}, discussed in Section~\ref{sec:rwdata}, for its broad range of content. This dataset contains 50 \SI{30}{\hertz} traces of 10 minutes each, each gathered by another user. Without further hyperparameter tuning, we applied the same process to generate a synthetic dataset as above. Figure~\ref{fig:bonus} summarises all metrics. These results again indicate a close match between the source and synthetic datasets. The under-representation of very-low-motion samples again occurs, but no other issues appear. We consider this as a strong indication of the generality of this approach. Intuitively, this is unsurprising, as the TimeGAN had no knowledge of any of our metrics. The only feedback available to the generator is the discriminator's loss. Without any additional steering, the generator learned from only this information source to match the original dataset very well.
\subsection{Discussion}
Based on the analysis above, we discuss the utility of the TimeGAN, compared to the FFT on several fronts.
\subsubsection{Dataset-specific manual design}: with the FFT, an expert needs to analyse the dataset, determining models for the distributions within the time series. In \cite{FFTmodel}, it is claimed that the model is general, as it can be adapted to another dataset by simply changing the parameters of the distributions. We note that this may not necessarily be the case. As the dataset was recorded in an indoor virtual experience, the yaw is approximated with a multi-modal Gaussian distribution, with their means representing the directions towards the room's walls. We do not expect this distribution to apply well to an outdoor virtual environment, where the user would likely look in any direction with similar probability. In applying TimeGAN however, no such dataset-specific design was applied. The model converts the source dataset to have a normal distribution, meaning it is insensitive to the distributions within the source dataset. The only type of tuning needed was hyperparameter tuning. During this process, we observed that the quality of the synthetic dataset was not particularly sensitive to the hyperparameters. Furthermore, no additional tuning was needed for the second dataset in Section~\ref{sec:generality}. Determining the optimal dataset from the different snapshots does require manual intervention, however this essentially comes down to comparing the plots of the metrics presented above, which can be performed rapidly by an individual with no expert knowledge, and could even be automated to an extent.
\subsubsection{Runtime}: the FFT is relatively fast, requiring only minutes of computation on a regular workstation computer. \glspl{GAN} however are notoriously computationally inefficient. Training our model on a workstation computer may take over 10 hours. We do note that this high runtime is not prohibitive for this application. Training needs to occur only once, after which the model can output new sequences in a matter of seconds.

Overall, we are convinced that this approach is capable of producing a large array of highly usable head rotation samples regardless of the specific head rotation distribution, for applications such as proactive viewport-dependent streaming and \gls{XR} beamforming.

\section{Conclusions}\label{sec:conclusions}
In this paper, we presented a novel approach for generating synthetic head rotation data for Extended Reality applications. We showed that, unlike the only other approach currently described in the literature, our TimeGAN-based approach is able to generate realistic data according to a range of metrics which together characterise the head rotation data. Our metrics incorporate where the users look and how they turn their heads. We expect this approach to be valuable to researchers in several \gls{XR}-related fields, including dynamic multimedia encoding and millimetre-wave beamforming. As such, we commit to releasing an open-source implementation of the system by this paper's publication date. In future work, we intend to reduce TimeGAN's tendency to generate datasets where very-low-motion samples are underrepresented in case of a wide motion distribution.

\begin{acks}
  Jakob Struye is funded by the Research Foundation Flanders (FWO), grant nr. 1SB0719N. (Part of) this research was funded by the ICON project INTERACT, realized in collaboration with imec, with project support from VLAIO (Flanders Innovation and Entrepreneurship). Project partners are imec, Rhinox, Pharrowtech, Dekimo and TEO. This research is partially funded by the FWO WaveVR project (Grant number: G034322N). 
\end{acks}
\bibliographystyle{ACM-Reference-Format}
\bibliography{timegan_headrots}

\end{document}